\documentclass[10pt]{article} 

\usepackage[preprint]{rlj} 

\usepackage{amssymb}            
\usepackage{mathtools}          
\usepackage{mathrsfs}           
\usepackage{graphicx}           
\usepackage{subcaption}         
\usepackage[space]{grffile}     
\usepackage{url}                
\usepackage{lipsum}    
\usepackage{subcaption}
\usepackage{svg}
\usepackage[textsize=tiny]{todonotes}

\title{Multi-Task Reinforcement Learning Enables Parameter Scaling}

\setrunningtitle{MTRL Enables Parameter Scaling}

\author{Reginald McLean\textsuperscript{1,2,$*$}, Evangelos Chatzaroulas\textsuperscript{3,$*$},  J.K. Terry\textsuperscript{2},  
Isaac Woungang\textsuperscript{1,$\dagger$}, Nariman Farsad\textsuperscript{1,$\dagger$}, Pablo Samuel Castro\textsuperscript{4,5,$\dagger$}}

\emails{reginald.mclean@torontomu.ca}

\affiliations{
$^{1}$\textbf{Department of Computer Science, Toronto Metropolitan University, Canada}\\
$^{2}$\textbf{Farama Foundation}\\
$^{3}$\textbf{Department of Computer Science, University of Surrey, United Kingdom}\\
$^{4}$\textbf{Mila - Québec AI Institute}\\
$^{5}$\textbf{Department of Computer Science and Operations Research, Université de Montréal, Canada}\\

\par 
$^*$ indicates equal contribution, $\dagger$ indicates equal advising
}

\contribution{
    We perform a thorough empirical evaluation that shows that parameter scaling with simple architectures exceeds performance of complex MTRL-specific architectures on Meta-World benchmarks.
    }
    {
    Increasing the scale of reinforcement learning models has been of particular interest of late \citep{pmlr-v235-obando-ceron24b, nauman2024bigger, schwarzer_bigger_2023}. This interest is likely due to the advances that other areas of research, such as computer vision \citep{DBLP:conf/cvpr/Zhai0HB22} and natural language processing \citep{Kaplan2020ScalingLF}, have made thanks to increasing the amount of data, compute, and model parameters during training. While some works have noted that na\"{i}vely scaling can be detrimental to performance \citep{pmlr-v235-obando-ceron24b}, MTRL models have been quietly increasing in size. We compare results across recent multi-task reinforcement learning (MTRL) specific architectures, finding that increasing the parameter count on a simple feed forward baseline architecture outperforms the MTRL specific architectures. 
    }

\contribution{
    We uncover a significant inverse relationship between task diversity and plasticity loss, showing that models trained on more tasks maintain neuron activation even as parameter counts increase.
    }
    {
    When increasing the parameter scale alone, recent works have found that networks exhibit a loss of network plasticity \citep{pmlr-v202-lyle23b}, or even a significant drop in performance when na\"{i}vely scaling \citep{pmlr-v235-obando-ceron24b}. Across three different sizes of benchmark tasks, ten, twenty-five, and fifty tasks, we uncover a previously unknown relationship between the number of tasks and the number of parameters. We find that by increasing both model scale and the number of tasks being trained on can be an effective method to mitigate plasticity loss.
    }

\contribution{
    We empirically identify that critic scaling provides greater benefits than actor scaling in MTRL settings.
    }
    {
    Recently, \citet{nauman2024bigger} uncovered that the performance of the critic of an actor-critic algorithm was more affected by scale in single task settings. We extend this question to MTRL and have similar findings: that the scaling of the critic is more tightly coupled with the performance of the MTRL agent as a whole than the scale of the actor. 
    }


\summary{Multi-task reinforcement learning (MTRL) aims to endow a single agent with the ability to perform well on multiple tasks. Recent works have focused on developing novel sophisticated architectures to improve performance, often resulting in larger models; it is unclear, however, whether the performance gains are a consequence of the architecture design itself or the extra parameters. We argue that gains are mostly due to scale by demonstrating that na\"{i}vely scaling up a simple MTRL baseline to match parameter counts outperforms the more sophisticated architectures, and these gains benefit most from scaling the critic over the actor. Additionally, we explore the training stability advantages that come with task diversity, demonstrating that increasing the number of tasks can help mitigate plasticity loss. Our findings suggest that MTRL's simultaneous training across multiple tasks provides a natural framework for beneficial parameter scaling in reinforcement learning, challenging the need for complex architectural innovations.}

\begin{document}

\maketitle  

\begin{abstract}
Multi-task reinforcement learning (MTRL) aims to endow a single agent with the ability to perform well on multiple tasks. Recent works have focused on developing novel sophisticated architectures to improve performance, often resulting in larger models; it is unclear, however, whether the performance gains are a consequence of the architecture design itself or the extra parameters. We argue that gains are mostly due to scale by demonstrating that na\"{i}vely scaling up a simple MTRL baseline to match parameter counts outperforms the more sophisticated architectures, and these gains benefit most from scaling the critic over the actor. Additionally, we explore the training stability advantages that come with task diversity, demonstrating that increasing the number of tasks can help mitigate plasticity loss. Our findings suggest that MTRL's simultaneous training across multiple tasks provides a natural framework for beneficial parameter scaling in reinforcement learning, challenging the need for complex architectural innovations.
\end{abstract}

\section{Introduction}
Reinforcement learning (RL) has emerged as the go-to method for super-human decision making in the past few years, achieving incredible results in Atari games \citep{schwarzer_bigger_2023}, the control of stratospheric balloons \citep{bellemare_autonomous_2020}, and the magnetic control of tokomak plasma for nuclear fusion  \citep{degrave_magnetic_2022}. The rise of RL has been in large part thanks to the introduction of neural networks for modeling actions and/or values from learning on many iterations of trial and error. During this process, it is possible for the RL agent to learn to utilize and exploit specific features to be successful. One of the drawbacks of this exploitation is that the RL agent can be limited in the number of tasks that it can accomplish, as the learned features may not generalize to other tasks.

\begin{figure*}[h!]
    \begin{subfigure}[]{0.5\textwidth}
        \centering
        \includegraphics[width=\linewidth]{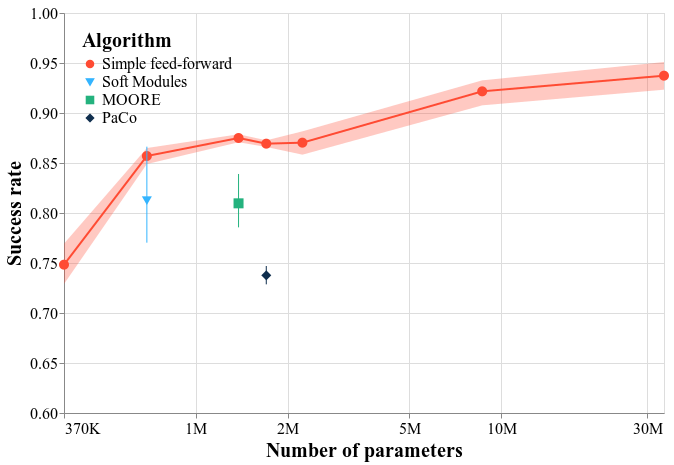}
        \caption{}
    \end{subfigure}%
    ~
    \begin{subfigure}[]{0.5\textwidth}
        \centering
         \includegraphics[width=\linewidth]{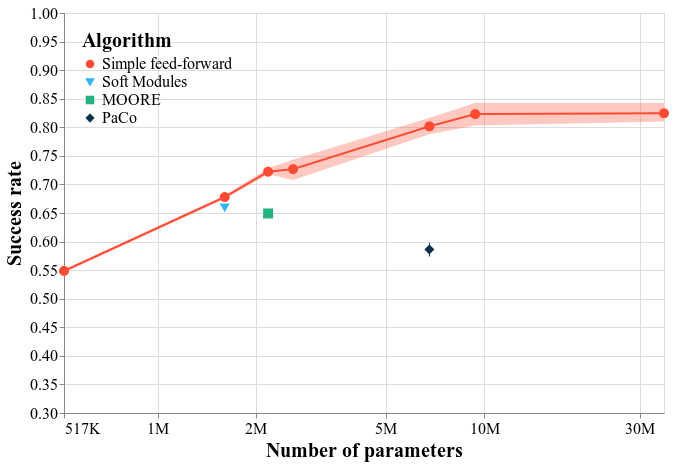}
        \caption{}
    \end{subfigure}

    \caption{Effects of scaling the number of parameters of actor and critic in multi-task, multi-head, soft actor-critic \citep{pmlr-v100-yu20a}, in the Meta-World (a) ten task and (b) fifty task benchmarks over ten random seeds. Y-axes report the inter-quartile mean. Shaded regions, for plots, and bars, for singular points, denote standard errors. MTRL specific methods are plotted here as singular points, while the Simple FF method is the result scaling a simple baseline to a similar number of parameters.}
    \label{fig:scaling_fig1}
\end{figure*}


To overcome this, RL researchers have developed methods of training multi-task reinforcement learning (MTRL) agents. These agents learn to accomplish multiple tasks using only a single policy and value function, with the aim of being able to bootstrap learning in one task by simultaneously learning on related tasks. In order to evaluate MTRL agents, researchers often utilize the Meta-World suite \citep{pmlr-v100-yu20a}, which provides two sets of benchmarks with different difficulty levels based on the number of tasks. Meta-World's unified observation and action space, along with its consistent robot workspace, makes it particularly suitable for MTRL research.

Researchers have approached MTRL through two main directions: gradient projection methods to limit interference between tasks \citep{10.5555/3495724.3496213}, and MTRL-specific architectures designed with human intuition about how agents should solve multiple tasks. While recent work \citep{kurin2022in} has shown that gradient projection methods do not significantly outperform well-regularized baselines, MTRL-specific architectures continue to push performance boundaries. However, these architectural innovations often lack comparison to simple baselines with similar parameter counts.

Recent works in computer vision \citep{DBLP:conf/cvpr/Zhai0HB22}, and language modeling \citep{Kaplan2020ScalingLF} have found that increasing the amount of compute to train the model, whether it is more parameters or more gradient steps, leads to models that perform better at their respective tasks. There may even be some signs of emergent capabilities at these scales \citep{wei2022emergent}. However, scaling in RL has proven more challenging, with research showing that na\"{i}vely increasing model parameters can actually harm performance \citep{pmlr-v235-obando-ceron24b}. Successful scaling in RL requires specific adaptations, such as modified network architectures \citep{pmlr-v235-obando-ceron24b}, policy and value function regularization \citep{nauman2024bigger}, or limiting parameter increases to world models only \citep{hansen2024tdmpc, DBLP:journals/corr/abs-2301-04104}.

Similar to recent attempts to scale model size in single-task RL that utilize architectural changes such as Soft Mixture of Experts \citep{pmlr-v235-obando-ceron24b}, or skip connections with regularization \citep{nauman2024bigger}, recent MTRL works have been able to leverage architectures to scale their models. However, there is not much insight into whether the MTRL architecture or the number of parameters provides the increase in performance. In order to investigate the correlation between model performance and scale, we first implement and benchmark four recent MTRL architectures on two Meta-World benchmarks. We demonstrate that a simple MTRL baseline, scaled to match these architectures' parameter counts, surpasses their performance. Further scaling of this basic architecture yields additional performance gains, as shown in Figure \ref{fig:scaling_fig1}.

Beyond raw performance improvements, we investigate whether both policy and value functions benefit equally from increased scale. Our analysis reveals that the scaling of the critic is directly related to MTRL performance. We also examine how scaling affects model plasticity - the ability to continually learn. Interestingly, we find that large models trained on few tasks show signs of reduced plasticity, while similarly sized models trained on more tasks maintain their adaptability. These findings suggest an important relationship between model capacity and task diversity in MTRL.

Based on these findings, the contributions of our work include:
\begin{itemize}
    \item Empirical evaluation that shows that parameter scaling with simple architectures \textbf{exceeds} performance of complex MTRL-specific architectures on Meta-World benchmarks.
    \item We empirically identify that critic scaling provides greater benefits than actor scaling in MTRL settings.
    \item We discover a previously unknown relationship between model size and task count: plasticity loss in large models can be mitigated by increasing the number of training tasks.
    \item An open source code base, publicly available with the final version, that can be used to further advance MTRL research.
\end{itemize}

\section{Related Work}
Before the introduction of a unified MTRL benchmark, single task RL benchmarks such as the Arcade Learning Environment (ALE) \citep{10.5555/2566972.2566979} were leveraged as a platform for MTRL research. However, using the ALE can be difficult due to large visual differences between games which can lead to negative interference \citep{pmlr-v100-yu20a}. Thus the introduction of a unified benchmark specifically for MTRL research was needed, such as Meta-World. Meta-World utilizes a unified observation and action space, and a single robot \& workspace, which allows for effective multi-task and meta reinforcement learning research. 

Upon publication, \citet{pmlr-v100-yu20a} produced an extension to Soft Actor-Critic (SAC) \citep{haarnoja2018soft}, specifically for MTRL. This extension, multi-task multi-head SAC (MTMHSAC), maintains $N$ independent entropy heads \& utilizes independent action heads for each task. The next MTRL architecture to be introduced was the Soft-Modularization (SM) architecture \citep{yang_multi_task_soft_mod}. This SM architecture posited that it is unclear whether gradients from all tasks should affect all parameters and which parameters should be shared. Thus to overcome this issue, SM proposed a soft-routing method. The \textbf{Pa}rameter-\textbf{Co}mpositional (PaCo) MTRL architecture creates a shared base representation and task-specific compositional vectors \citep{sun2022paco}. For each task, parameters are generated by combining the shared base representation with the corresponding task-specific vector. This design ensures that the shared components learn from all tasks while maintaining task-specific adaptability through the compositional vectors. The architecture explicitly separates gradient flow between shared and task-specific components, with shared parameters updated using all task losses while task-specific components only receive gradients from their respective tasks. Lastly, a \textbf{M}ixture \textbf{O}f \textbf{OR}thogonal \textbf{E}xperts (MOORE) was proposed in \citet{hendawy2024multitask}. In this architecture a mixture of experts processes the incoming state into a shared representation space, which is then used in combination with task specific information to generate task relevant representations. 

Recent works in RL have attempted to scale the number of parameters to improve performance of the RL agent. One of the first attempts to do this was a model-based approach that showed it was possible to learn a policy entirely in a world model \citep{Kaiser2020Model}. Building upon that work, there have been multiple attempts to scale models when using \citet{Kaiser2020Model}'s Atari 100k dataset, including representation learning approaches \citep{schwarzer2021dataefficient, farebrother2023protovalue, schwarzer_bigger_2023}, soft mixture-of-experts \citep{pmlr-v235-obando-ceron24b}, and regularization based \citep{nauman2024bigger}\footnote{Appendix E of \citet{nauman2024bigger} shows some signs that scaling also aids in MTRL}. Recent work in model-based RL have shown that scaling the world model achieves significant improvements \citep{DBLP:journals/corr/abs-2301-04104, hansen2024tdmpc, wang2024efficientzero} with the caveat that most of the scaled parameters are in the world model itself, not the RL agent. Recent works have also shown that na\"{i}vely scaling parameters can be detrimental to overall performance of the RL agent \citep{pmlr-v235-obando-ceron24b}. 

Recently, the single task reinforcement learning has seen an influx of retrospective works that have shaped our approach to this work. This line of work began with a study into the implementation details of Proximal Policy Optimization and Trust-Region Policy optimization algorithms \citep{Engstrom2020Implementation}, revealing how subtle implementation choices can significantly impact performance. These findings were further extended into a large scale study into what matters in on-policy RL \citep{andrychowicz2021what}, while \citet{taiga2023investigating} demonstrated how pre-training on simple game variants affects the generalization abilities of RL agents of various sizes. Recent works have also investigated why RL agents lose their network plasticity and have proposed several mitigation techniques including: loss landscape stabilization \citep{pmlr-v202-lyle23b}, plasticity injection \citep{nikishin2023deep}, neuron recycling \citep{pmlr-v202-sokar23a}, and appropriate regularization \citep{10.5555/3692070.3693586}. These plasticity findings are particularly relevant to MTRL, where agents must maintain flexibility to learn multiple tasks simultaneously. Other works have investigated the effects and consistency of hyper-parameter selections \citep{ceron2024on}, which informed our systematic approach to evaluating MTRL architectures across different configurations and settings.

While these architectural innovations and optimization techniques have advanced MTRL, our work suggests that simpler approaches focusing on parameter scaling and critic accuracy may be equally or more effective. This aligns with recent trends in other areas of deep learning where scale has proven to be a crucial factor in performance improvements. Our findings indicate that the field's focus on complex architectural solutions may be better directed toward ensuring robust critic training and adequate model capacity.

\section{Background}
Reinforcement learning attempts to provide solutions to sequential decision making problems where an agent interacts with an environment during a set of discrete time steps $t = 0,1,2,3,...$. During each time step $t$, the agent receives a representation of an environment's state $s_t$, selects an action $a_t$, and applies that action to the environment. As a consequence of applying this action, the agent then receives a new state $s_{t+1}$, and a scalar reward $r_{t+1}$ for transitioning from state $s_t$ to $s_{t+1}$. This framework is formalized as a Markov Decision Process (MDP), represented as $(\mathcal{S}, \mathcal{A}, \mathcal{P}, \mathcal{R}, \gamma, \rho)$ with the set of all possible states $\mathcal{S}$, the set of available actions $\mathcal{A}$, $\mathcal{P} : \mathcal{S} \times \mathcal{A} \rightarrow \mathcal{S}$ defines the transition function, $\mathcal{R} : \mathcal{S} \times \mathcal{A} \rightarrow \mathbf{R}$ is the reward function, $\gamma \in [0, 1)$ is the discount factor, and $\rho$ is the initial state distribution. In order to select actions, an agent's behavior is formalized using a policy $\pi : \mathcal{S} \rightarrow \mathcal{A}$. The objective of the agent then is to maximize the expected sum of returns through interactions with the environment according to policy $\pi$ as $\mathbf{E}_{(s_t,a_t)\sim \mathcal{P_\pi}} [R(\tau)]$ where $\mathcal{P_\pi}$ is the transition distribution induced by policy $\pi$. 

The MTRL problem extends the usage of an MDP across all $N$ tasks the agent must interact with. We assume that task $i$ is drawn from a distribution of tasks $p(\mathcal{T})$. Instead of a single MDP, the MTRL agent must interact with MDPs $(\mathcal{S}, \mathcal{A}, \mathcal{P}_i, \mathcal{R}_i, \gamma, \rho_i)$ where the transition function $\mathcal{P}_i$, and reward function $\mathcal{R}_i$ are specific to sampled task $i$. The goal of the agent is to learn a single, task conditioned, policy $\pi(a_t|s_t, z)$ where $z$ is a representation of which task $i$ the agent is interacting with. The objective of the agent is now to maximize expected returns across all $N$ tasks as $\mathbf{E}_{i\sim p(\mathcal{T})}[\mathbf{E}_{(s_t,a_t)\sim \mathcal{P_\pi}} [R_i(\tau)]]$.  In this work we leverage Soft Actor-Critic \citep{haarnoja2018soft}, as the underlying RL algorithm to maximize the expected returns.

\section{Experimental Methodology}
This section outlines the experimental setup of our study. We aim to answer the following questions:
\begin{enumerate}
    \item Does increasing the number of parameters account for the recent advances in MTRL?
    \item Does increasing the number of parameters predictably increase MTRL performance?
    \item Where does scaling aid in improving performance the most?
    \item How does scaling the number of parameters affect plasticity loss? Does this effect change with different number of tasks?
\end{enumerate}

\subsection{Benchmark environments}
\begin{figure*}[h]
    \begin{subfigure}[]{0.24\textwidth}
        \centering
        \includegraphics[width=1\linewidth]{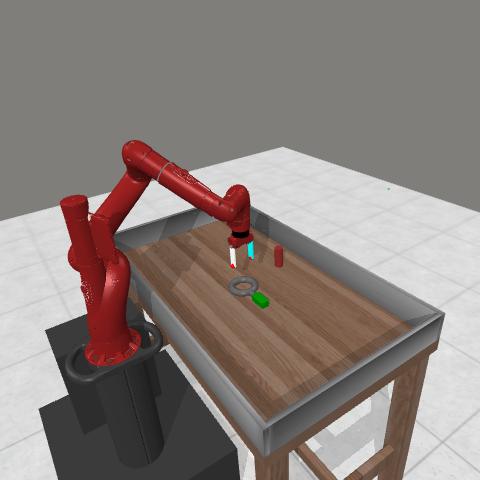}
        \caption{}
    \end{subfigure}%
    ~
    \begin{subfigure}[]{0.24\textwidth}
        \centering
        \includegraphics[width=1\linewidth]{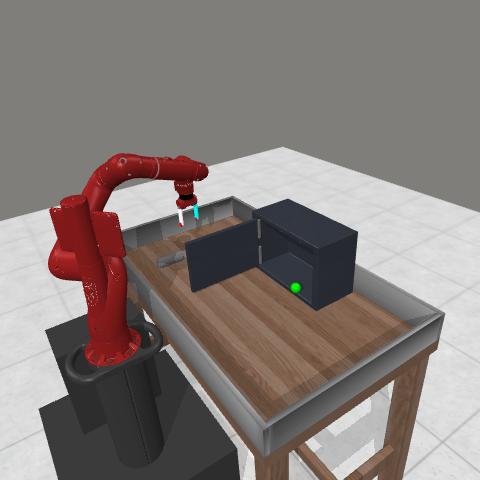}
        \caption{}
    \end{subfigure}
    ~
    \begin{subfigure}[]{0.24\textwidth}
        \centering
        \includegraphics[width=1\linewidth]{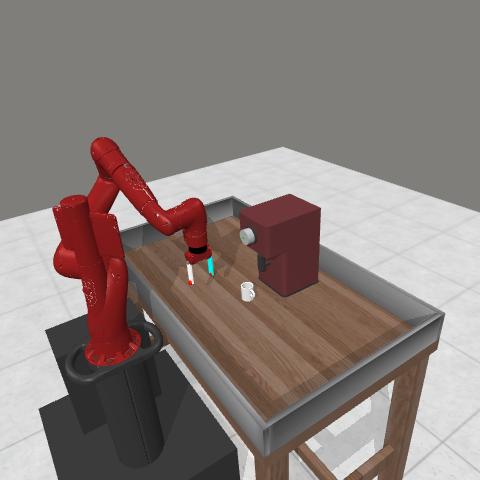}
        \caption{}
    \end{subfigure}%
    ~
    \begin{subfigure}[]{0.24\textwidth}
        \centering
        \includegraphics[width=1\linewidth]{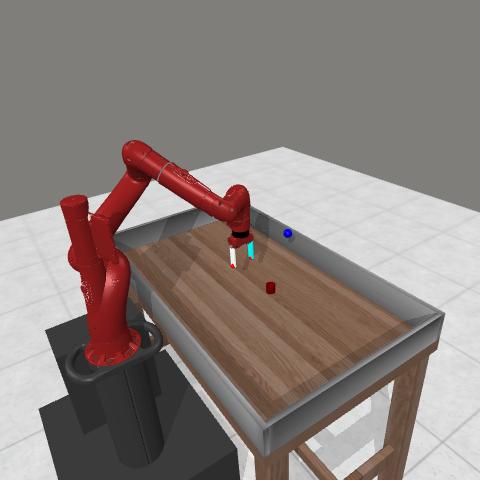}
        \caption{}
    \end{subfigure}
    \caption{Sample tasks from Meta-World. (a) assembly, where the MTRL agent must grasp a wrench and place it on a peg. (b) door close, where the agent must swing the door from the current position to the green goal to close it. (c) coffee button, where the agent must press a button on the front of the coffee machine. (d) pick place, where the agent must grasp the red object and move it to the blue goal. }
    \label{fig:metaworld_tasks}
\end{figure*}

We focus on the two available benchmarks from Meta-World \citep{pmlr-v100-yu20a}. The first benchmark contains 10 tasks, we will refer to it as MT10. The second benchmark contains fifty tasks, which we will refer to as MT50. In Figure \ref{fig:metaworld_tasks} we show some of the tasks from Meta-World. We refer the interested reader to \citet{pmlr-v100-yu20a} for the remaining tasks. Each state vector has a one-hot task ID vector appended to it, indicating which task the state vector belongs to. We train our MTRL agent in MT10 for twenty million timesteps and one-hundred million timesteps for MT50, with two million timesteps per task. We utilize per-task replay buffers, with an equal number of samples per task for each actor \& critic update. We evaluate our policies for fifty episodes, every forty episodes of training. For MT10 this equates to every 200,000 timesteps while MT50 is evaluated every 1,000,000 timesteps. For each experiment we report results across 10 random seeds. Unless otherwise noted, we report the inter-quartile means (IQM) following recommendations from \citet{agarwal2021deep}. 

\subsection{Architectures}
We implement four MTRL architectures from the MTRL literature: multi-task, multi-head, Soft Actor-Critic (MTMHSAC) \citep{pmlr-v100-yu20a}, MTRL with Soft Modularization (SM) \citep{yang_multi_task_soft_mod}, MTRL with Mixture of Orthogonal Experts (MOORE) \citep{hendawy2024multitask}, and Parameter Compositional MTRL (PaCo) \citep{sun2022paco}. These architectures were selected to represent the evolution and diversity of MTRL approaches, ranging from fundamental baseline methods, MTMHSAC, to sophisticated state-of-the-art architectures like MOORE. Each architecture embodies different design philosophies in handling multi-task learning, from MTMHSAC's simple feed forward design with separate action heads to MOORE's expert-based parameter sharing strategy. We include a diagram of the respective architectures in Figure \ref{fig:architectures}. In Supplementary Materials \ref{sec:baselines} we report the baseline results of our implemented architectures in comparison to the published results.

In addition to their architectural contributions, SM and PaCo incorporate additional optimization tweaks beyond their architectural innovations. To isolate and evaluate the impact of architectural choices alone, we implement these methods without their supplementary losses. This focused comparison necessarily means our results may differ from those reported in the original papers. Furthermore, SM's published results used an earlier version of Meta-World's reward functions, making direct numerical comparisons to their reported results inappropriate.

\begin{figure}[t]
    \centering
    \includegraphics[width=\linewidth]{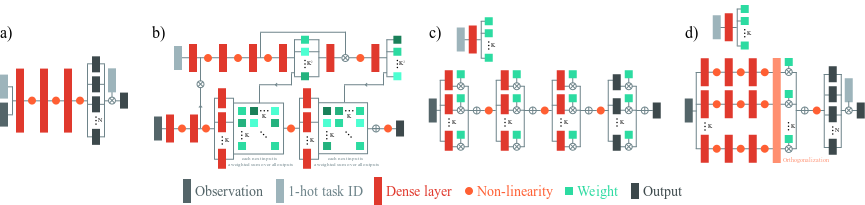}
    \caption{Visualization of architectures used in this work. (a) MTMHSAC from \citet{pmlr-v100-yu20a}, (b) Soft-Modularization from \citet{yang_multi_task_soft_mod}, (c) PaCo from \citet{sun2022paco}, and (d) MOORE from \cite{hendawy2024multitask}.}
    \label{fig:architectures}
\end{figure}

\subsection{Hyperparameters}
To ensure fair comparisons across architectures, we standardize a set of common MTRL hyperparameters including batch size, log standard deviation bounds, and replay buffer capacity. While some of our chosen values differ from those in the original publications, this standardization is essential for meaningful comparative analysis. For instance, we adopt a minimum log standard deviation of $e^{-20}$ across all methods, in contrast to MOORE's original $e^{-10}$, and implement a uniform replay buffer size of $100k \times N$ (where N is the number of tasks) rather than SM's larger $200k \times N$ buffer. These modifications enable direct architectural comparisons by eliminating performance variations that might stem from differing hyperparameter choices. 


\section{Empirical Results}

\subsection{Comparison with MTRL Architectures}
Figure \ref{fig:scaling_fig1} reports results on the MT10 and MT50 benchmarks across various parameter scales. In order to investigate scaling the number of parameters in our Actor Critic algorithm, we first choose parameter scales that match the number of parameters in each of our selected baselines, SM, MOORE, and PaCo. In Figure \ref{fig:scaling_fig1} we show the IQM plots of using a simple feed-forward neural network architecture combined with increasing the number of parameters to match our selected baselines. To do so we use 3 hidden layers\footnote{We also explored using 4 hidden layers and found that performance increased in comparison to the MTRL architecture baselines, but provided no gain compared to the 3 layer depth networks. We include those results in Supplementary Material \ref{sec:depth_4_results}. We hypothesize that going to four or more layers may require architecture changes, such as skip connections \cite{7780459}, in order to exceed three layer performance due to gradient vanishing.} for our actor and critic, and increase the width of each neural network layer until the number of parameters in the actor and critic is, at most, equal to the parameter count of the MTRL specific architecture.  As Figure \ref{fig:scaling_fig1} shows, we find that simply scaling the number of parameters to match the number of parameters in each baseline matches or, exceeds baseline architecture performance. These results confirm our initial hypothesis that scaling accounts for recent MTRL advances.

\begin{figure*}[h!]
    \centering
    \begin{subfigure}[t]{0.5\textwidth}
        \centering
        \includegraphics[width=\linewidth]{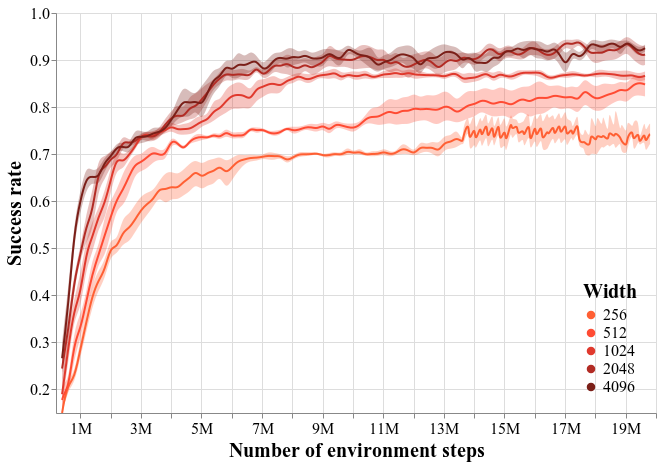}
        \caption{}
        \label{fig:scaling_mt10}
    \end{subfigure}%
    ~
    \begin{subfigure}[t]{0.5\textwidth}
        \centering
        \includegraphics[width=\linewidth]{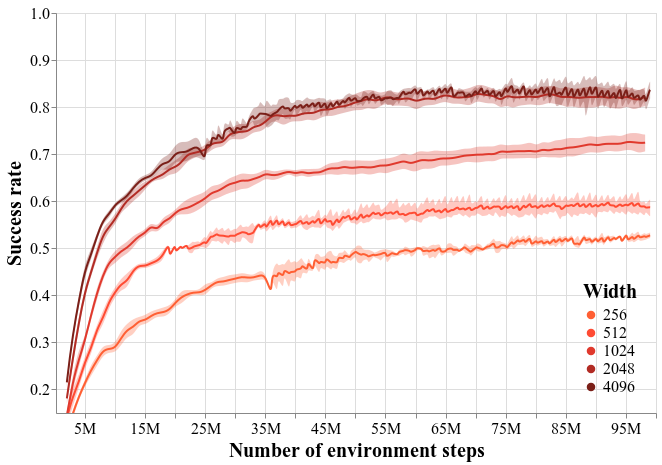}
        \caption{}
        \label{fig:scaling_mt50}
    \end{subfigure}
    \caption{Extended Scaling Results. Each line is the average IQM produced by a feed-forward network scaled to a certain parameter count, with shaded regions indicating 95\% CIs. Colours indicate which parameter count the model used. (a) MT10 scaling results, (b) MT50 scaling results.}
    \label{fig:further_scaling}
\end{figure*}

\subsubsection{Increasing Scale}
To further explore the effect that scaling has on MTRL performance in Meta-World, we continue experimenting with increasing the parameter counts of our models. In Figure \ref{fig:scaling_mt10} and \ref{fig:scaling_mt50} we report the IQM on various scales of models. We find that as we continue to scale, there are some benefits of making the models increasingly larger. However, there does seem to be diminishing returns from scaling for performance. 

\subsection{Who benefits the most from scale, the actor or critic?}
In our previous experiments we show that the effect of scaling the number of parameters can have great effects in both the MT10 and MT50 benchmarks. Now we seek to answer whether the actor or the critic benefits from the effects of scale more. To perform this experiment, we use the MT10 benchmark and select the network from the extended scaling experiments with the layer width of 1024 as our baseline. This model size provides a decent trade-off between performance and speed for these experiments. In order to determine whether the actor or critic benefits from scaling more, we increase the actor's layer width to 1024 while maintaining a critic width of 400, and vice versa for the actor. In addition to this, we further reduce the size of our models from a width of 400 to 200 to further limit the ability for the shrunk model to learn.

\begin{figure*}[t!]
    \centering
    \includegraphics[width=\linewidth]{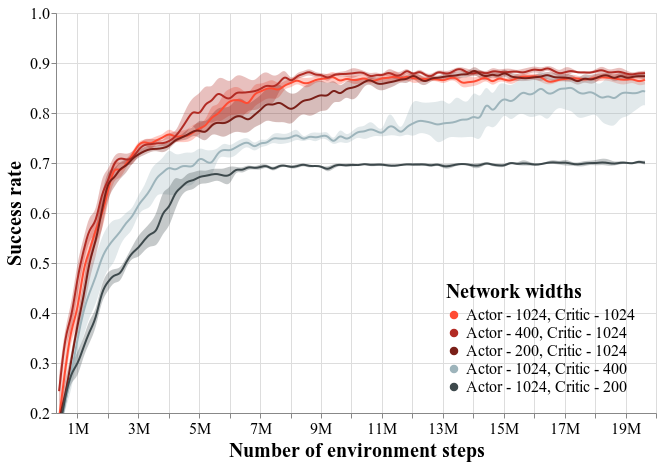}
    \caption{Where does scale matter the most? Here we report the IQM through training across various actor \& critic size configurations. Our baseline model uses equally sized actor \& critics using layer width 1024. We then iterate over various actor \& critic widths to determine which component benefits more from scale. These results show that the ability of the critic is more affected by the scaling of parameters, in line with results from \citet{nauman2024bigger} in the single task setting.}
    \label{fig:actor_vs_critic_scaling} 
\end{figure*}

Figure \ref{fig:actor_vs_critic_scaling} reveals a clear asymmetry in how scaling affects different components of MTRL architectures. When using our actor and critic both with width 1024 as baseline, we observe that experiments maintaining a critic of size 1024 consistently achieve comparable performance to this baseline, regardless of actor size reduction. Conversely, configurations with a full-sized actor (width 1024) but reduced critic (width 400 or 200) show substantial performance degradation. This asymmetric response to parameter reduction demonstrates that critics benefit more significantly from increased model capacity in multi-task settings. This finding aligns with our intuitions, as critics must approximate multiple state-action value functions across diverse tasks, thereby requiring greater representational capacity. Our results extend and confirm similar observations from single-task RL settings reported by \cite{nauman2024bigger}, suggesting this asymmetry may be a fundamental property of actor-critic architectures rather than a task-specific phenomenon.

\subsection{Impact of Scale on Plasticity} \label{sec:scale_plasticity}
\begin{figure*}[h!]
    \centering
    \begin{subfigure}[t]{0.5\textwidth}
        \centering
        \includegraphics[width=\linewidth]{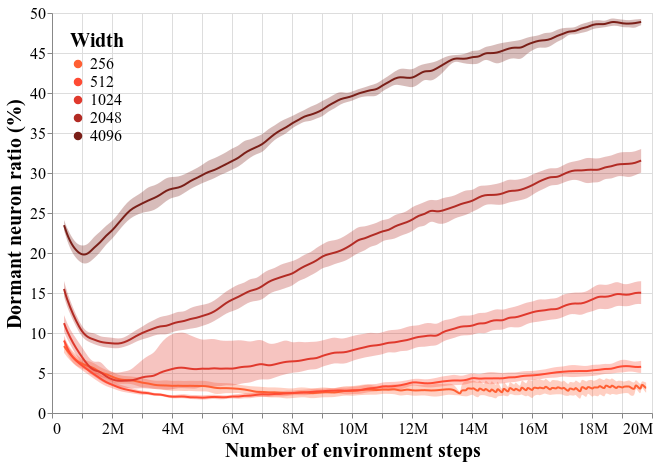}
        \caption{}
        \label{fig:}
    \end{subfigure}%
    ~
    \begin{subfigure}[t]{0.5\textwidth}
        \centering
        \includegraphics[width=\linewidth]{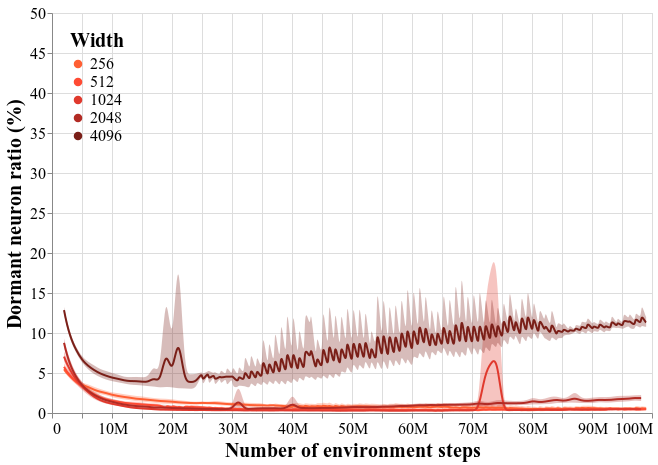}
        \caption{}
        \label{fig:}
    \end{subfigure}
    \caption{Impact of Scale on Plasticity. (a) MT10 (b) MT50. In MT10, we find that a bigger model has significantly more dormant neurons through training than smaller models. However, in MT50 we find that increasing the number of parameters and the number of tasks seems to limit plasticity loss. }
    \label{fig:scale_vs_plasticity}
\end{figure*}

Next we investigate the effects that scaling has on plasticity loss in our policy and value functions. We implement several metrics for approximating the loss of plasticity measure, similar to the prior work of \cite{10.5555/3692070.3693586}. We find that the percentage of dormant neurons to be the most representative metric, but we include plots for each of the remaining metrics in Supplementary Materials \ref{sec:plasticity_metrics}. In Figure \ref{fig:scale_vs_plasticity}, we report the percentage of dormant neurons when training on MT10 and MT50. When examining the MT10 dormant neurons, we find that as we scale the number of parameters we see an increase in the number of dormant neurons, with our highest parameter count model Width 4096 exhibiting the highest percentage of dormant neurons. This somewhat contrasts with our earlier results from scaling. While the performance in terms of success rate improves, there are signs of losing the ability to learn new tasks when training on MT10. When examining the MT50 dormant neuron percentages we find that at these parameter scales with fifty tasks, there is almost no indication of plasticity loss. This indicates that there is a relationship between the number of parameters in the actor \& critic, the number of tasks that the agent is interacting with, and the plasticity of the agent's networks.

This observation reveals a critical insight into the dynamics of plasticity in MTRL systems: the ratio between model capacity and task diversity appears to govern plasticity loss. In MT10, the limited task diversity relative to model size allows networks to specialize neurons for specific tasks, leaving portions of the network underutilized and potentially dormant. However, in MT50, the increased task diversity creates sufficient learning pressure to activate neurons throughout the network, even as model size increases. This finding suggests an important design principle for MTRL architectures—model scaling should be balanced with commensurate increases in task diversity to maintain neuronal activity and preserve plasticity. The absence of dormant neurons in large models trained on MT50 indicates that task diversity may serve as a natural regularizer against plasticity loss, enabling efficient utilization of increased parameter capacity without sacrificing the ability to learn new tasks.

\subsection{Task \& Parameter Scaling}
\begin{figure*}[h!]
    \centering
    \includegraphics[width=\linewidth]{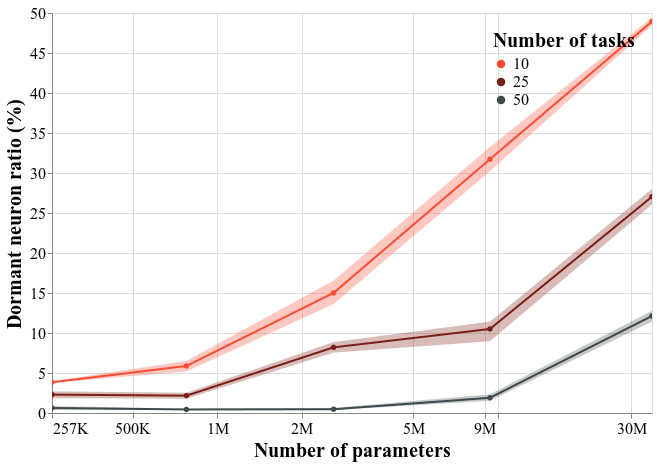}
    \caption{The effect of increasing the number of model parameters, when training on various numbers of tasks, on the percentage of dormant neurons in the MTRL agent. There seems to be some relationship between the number of tasks and the number of parameters, where scaling both leads to an effective usage of model parameters.}
    \label{fig:10_25_50_tasks}
\end{figure*}

To further examine the relation between the number of parameters, the number of tasks, and an agent's network plasticity, we perform the following experiments. First we create a new set of tasks called MT25 that contains the ten tasks from MT10, and the remaining tasks are a mix of solved and unsolved tasks from MT50. Next, we train MTRL agents across various scales of model sizes. Figure \ref{fig:10_25_50_tasks} shows the dormant neuron percentages plotted against the number of model parameters for the ten, twenty, and fifty tasks benchmark sets. From this experiment we find that large models trained on a large number of tasks exhibit much lower rates of dormant neurons, and less plasticity loss.

This finding has significant implications for scaling in RL. The inverse relationship between task diversity and plasticity loss suggests that scaling both model size and task count simultaneously may be a more effective strategy than focusing on either dimension alone. Rather than viewing large models as susceptible to plasticity loss, our results indicate that increasing task diversity creates sufficient learning pressure to maintain neuron activation across the network. This challenges the conventional wisdom that larger models necessarily suffer from greater plasticity issues and points toward a potential method for scaling RL models.

\section{Limitations}
While our analysis demonstrates that MTRL enables effective scaling of reinforcement learning models through multi-task training, several limitations warrant discussion. Our study primarily focuses on Meta-World, which, despite being a valuable MTRL benchmark, has inherent constraints (a) tasks are limited to short time horizons, (b) all tasks provide dense reward signals, (c) tasks operate in isolated environments with no occlusions, (d) MTRL agents receive privileged state information rather than pixel observations, (e) Meta-World tasks are much more similar than other RL tasks (i.e. tasks in the ALE \citep{10.5555/2566972.2566979}). Despite these limitations, we contend that our findings provide valuable insights applicable across different task domains and MTRL methods. The fundamental relationship between parameter scaling, task diversity, and performance likely extends beyond Meta-World's specific characteristics.


\section{Conclusion}
This work proposes that recent advances in multi-task reinforcement learning (MTRL) can be attributed to the number of parameters that recent works have used in their models and not the specialized MTRL architectures that each work proposes. In order to test this hypothesis, we implemented four MTRL architectures and re-benchmarked them across two Meta-World benchmarks. We then took a simple feed-forward baseline architecture and increased the number of parameters to match the number of parameters in the MTRL specific architectures. In doing so, we found that simply scaling the feed-forward neural network architecture outperformed the MTRL specific architectures. Next, we explored the effects of scaling the actor or the critic. We found that scaling the critic is more beneficial than scaling the actor, likely due to the Q-function having to fit multiple state-action functions. We also explored how far we can scale feed-forward neural networks to continue improving performance. While performing these experiments, we also discovered a previously unknown relationship between the number of tasks being trained on and the number of parameters in the model and plasticity: increasing both the number of parameters and the number of tasks being used can be an effective way to limit plasticity loss in RL. Finally, we acknowledged that Meta-World can be a simpler benchmark in comparison to other RL problems available in the literature but we hope that this work can provide a simple framework for future work exploring the scaling of reinforcement learning methods.  

Our results likely have implications for single task reinforcement learning. We have found that a simple plasticity loss mitigation strategy may be to increase the number of tasks that the reinforcement learning agent is currently being trained on. This may be because the additional tasks act as auxiliary tasks \citep{jaderberg2017reinforcement}. However, in some scenarios it may not be possible to add additional tasks for the agent to train on. In these cases, it may be possible to mitigate plasticity loss by increasing the number of parametric variations that the agent interacts with. In Meta-World, each task has up to fifty variations of object and/or goal location. It may in fact be these variations, in combination with a large number of tasks, that allows for better plasticity loss mitigation. Thus, it may be useful to investigate the introduction of smaller sub-tasks or parametric variations when attempting to scale single task reinforcement learning. Additionally, as we continued to scale our networks, we found that experiments would become more computationally expensive. It is likely a useful line of future work to explore how to leverage a relatively small number of parameters but have the performance of a much larger model when training online. 

\summary{}

\clearpage
\appendix


\bibliography{main}
\bibliographystyle{rlj}

\beginSupplementaryMaterials

\section{Additional MTRL Architectures}
Soft Modules:
\begin{enumerate}
    \item Smaller replay buffers, we use $100k * N$ where N is the number of tasks
    \item They apply an additional dynamic weighting scheme for summing the per-task losses, we utilize a uniform weighting of task losses
\end{enumerate}

MOORE:
\begin{enumerate}
    \item We use a minimum logarithm standard deviation output of $e^{-20}$ instead of $e^{-10}$
    \item The main text of \citet{hendawy2024multitask} suggests that they use 20 million timesteps for MT50, however the appendix suggests that they report results over 100 million timesteps. We explicitly use 100 million timesteps.
    \item We evaluate over fifty episodes, while \citet{hendawy2024multitask} reports performance over 10 evaluation episodes.
\end{enumerate}

PaCo:
\begin{enumerate}
    \item Their results leverage a weight reset and a loss maskout in order to stabilize training. We do not leverage those tricks
\end{enumerate}


\section{Depth 4}\label{sec:depth_4_results}

\begin{figure}[h!]
    \centering
    \includegraphics[width=\linewidth]{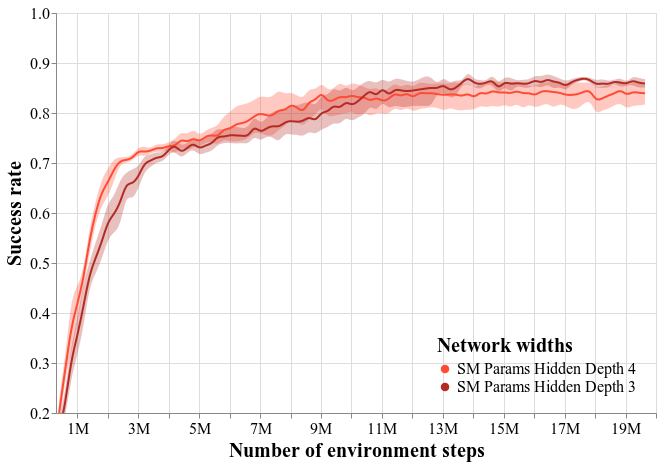}
    \caption{Effect of depth with SM number of parameters on MT10}
    \label{fig:enter-label}
\end{figure}

\begin{figure}[h!]
    \centering
    \includegraphics[width=\linewidth]{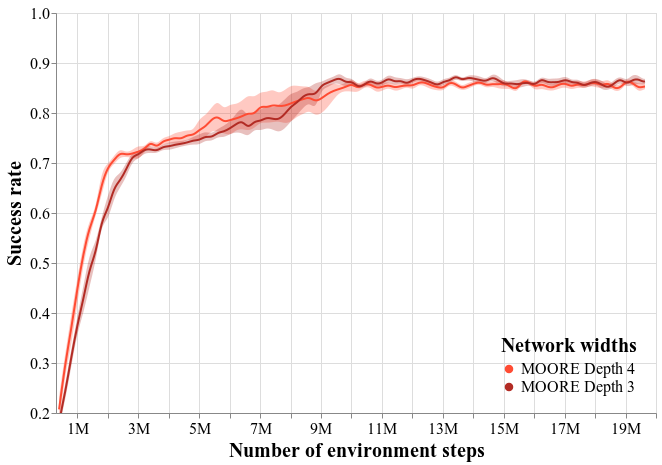}
    \caption{Effect of depth with MOORE number of parameters on MT10}
    \label{fig:enter-label}
\end{figure}

\begin{figure}[h!]
    \centering
    \includegraphics[width=\linewidth]{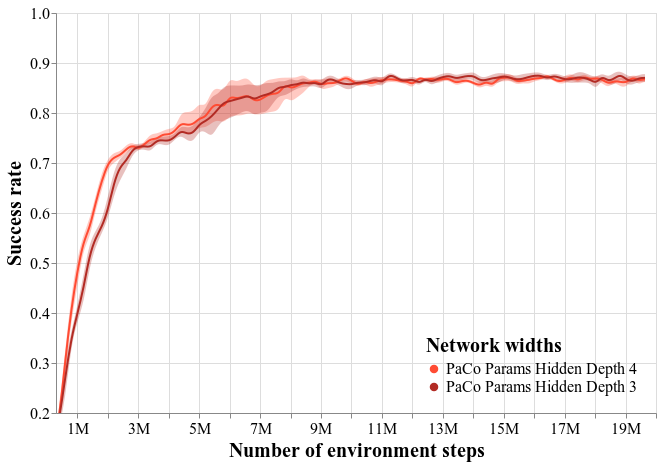}
    \caption{Effect of depth with PaCo number of parameters on MT10}
    \label{fig:enter-label}
\end{figure}

\section{Plasticity Metrics}\label{sec:plasticity_metrics}

\begin{figure}[h!]
    \centering
    \includegraphics[width=\linewidth]{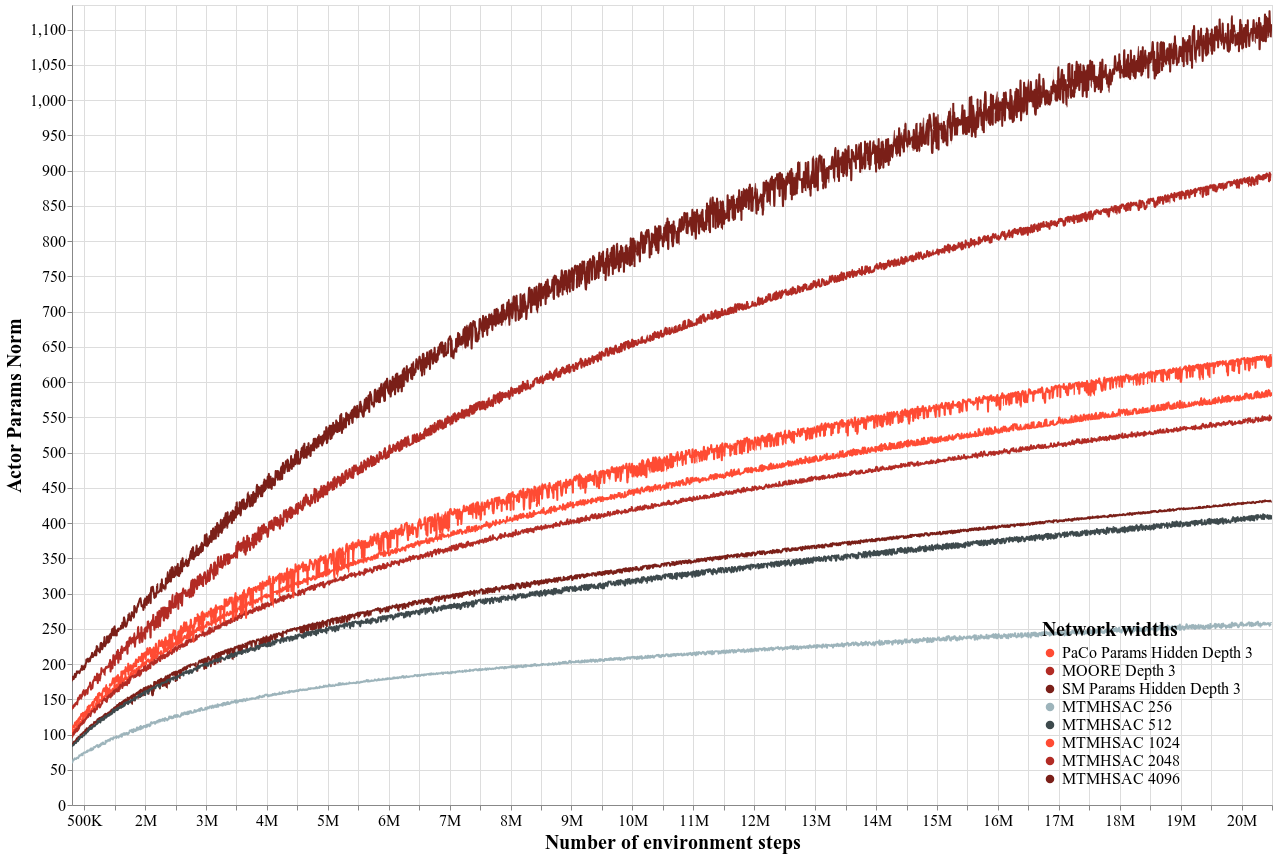}
    \caption{Actor Parameter Norms on MT10}
    \label{fig:enter-label}
\end{figure}

\begin{figure}[h!]
    \centering
    \includegraphics[width=\linewidth]{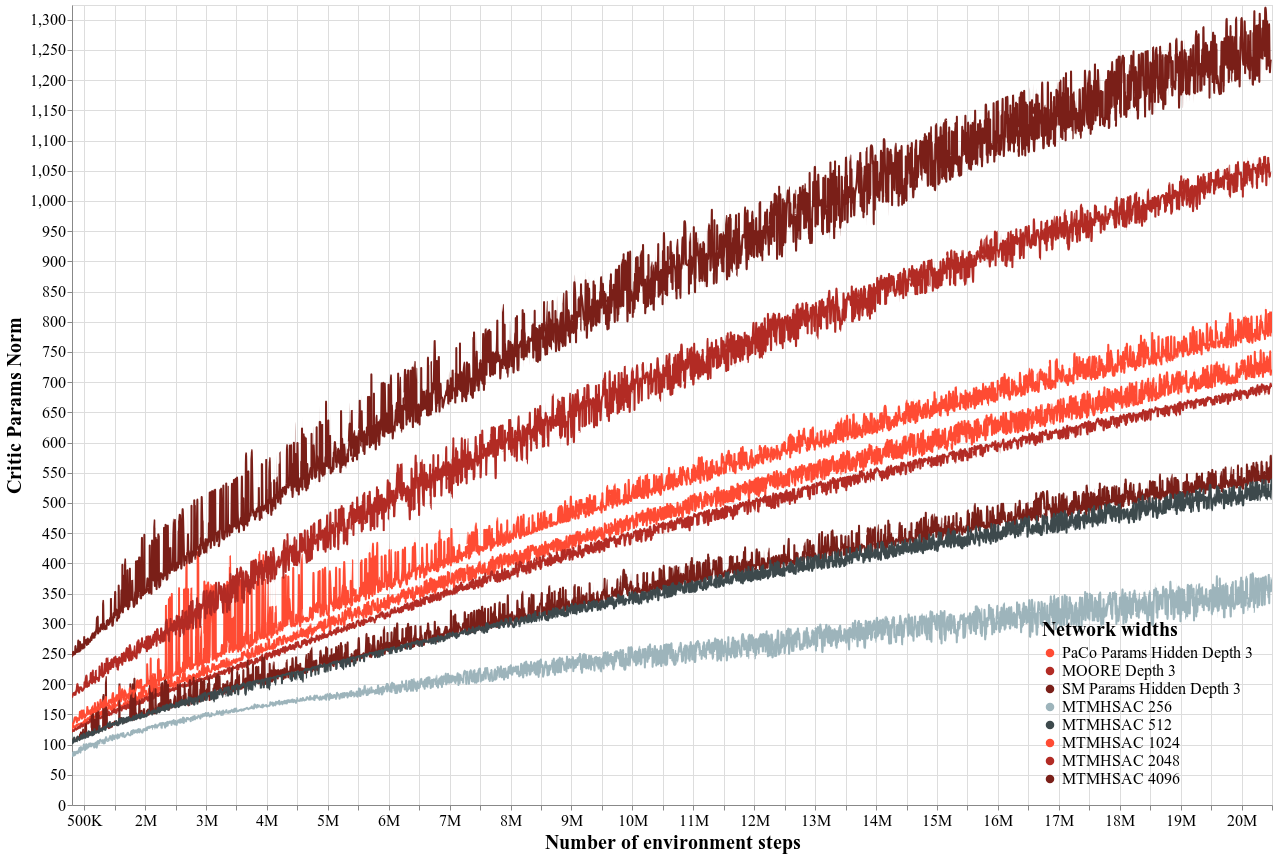}
    \caption{Critic Parameter Norms on MT10}
    \label{fig:enter-label}
\end{figure}

\section{Baseline Architecture Results} \label{sec:baselines}
In this section we report the results of our implemented MTRL architectures in relation to their publications. Results indicate that, even across various evaluation methods, that our implementations are correct. 

\begin{table}[h!]
    \centering
    \begin{tabular}{c|c|c|c|c}
        Architecture & MT10 (Ours) & MT10 (Pub) & MT50 (Ours) & MT50 (Pub) \\ \hline
        Soft Modularization \citep{yang_multi_task_soft_mod} & 71.4 & 71.8 & NA & NA \\
        PaCo \citep{sun2022paco} & 73.6 & 71.6 & 58.9 & 57.3 \\
        MOORE \citep{hendawy2024multitask} & 83.2 & 88.7 & 72.0 & 72.9 \\
    \end{tabular}
    \caption{Replication of selected MTRL results. For the Soft Modularization (SM) results, we note that SM reports results on an earlier version of Meta-World not currently available. For our reproduction we added the V1 rewards to the MT10 environments. For our PaCo MT10 results we compare to their "PaCo-Vanilla" results from Table 2 in \citet{sun2022paco}. This is the version of their architecture that does not include any loss masking or resets.}
    \label{tab:my_label}
\end{table}

\end{document}